\documentclass[conference]{IEEEtran}
\IEEEoverridecommandlockouts

\usepackage{cite}
\usepackage{amsmath,amssymb,amsfonts}
\usepackage{algorithmic}
\usepackage{graphicx}
\usepackage{textcomp}
\usepackage{xcolor}

\usepackage{multicol}
\usepackage{subcaption}

\def\BibTeX{{\rm B\kern-.05em{\sc i\kern-.025em b}\kern-.08em
    T\kern-.1667em\lower.7ex\hbox{E}\kern-.125emX}}

\begin{document}

\title{Cooking State Recognition from Images Using Inception Architecture}

\author{
Md Sirajus Salekin, Ahmad Babaeian Jelodar, and Rafsanjany Kushol
\thanks{*Corresponding contact email : \texttt{salekin@mail.usf.edu}}
}


\maketitle









\begin{abstract}
A kitchen robot properly needs to understand the cooking environment to continue any cooking activities. But object's state detection has not been researched well so far as like object detection. In this paper, we propose a deep learning approach to identify different cooking states from images for a kitchen robot. In our research, we investigate particularly the performance of Inception architecture and propose a modified architecture based on Inception model to classify different cooking states. The model is analyzed robustly in terms of different layers, and optimizers. Experimental results on a cooking dataset demonstrate that proposed model can be a potential solution to the cooking state recognition problem.
\end{abstract}

\begin{IEEEkeywords}
Cooking object classification, Cooking state recognition, Inception architecture, Kitchen robot, Object state classification
\end{IEEEkeywords}

\section{Introduction}
\IEEEPARstart{N}owadays scientists are being more interested to work with robots which can be involved in our day to day life. Already robots are playing a great role in different automated systems such as packaging, manufacturing products, finding out the faults of different productions etc. One of the recent research interest areas is how we can replace the human involvement in the kitchen by robots, especially in terms of cooking activities. For doing this, a robot needs to perceive and understand different cooking states while they are cooking. Although object detection from image got much improvement, unfortunately, there are not many significant works on different objects' state detection (for instance: in our case cooking states such as diced, juiced, sliced etc.). 

Robots are being used in the automation industry replacing human force. But those tasks are done in a very structured environment and robots do not need any other specific tools to do those. But if robots are involved in our day to day life activities such as cooking in the kitchen, they need to use a lot of instruments in an unstructured environment. For instance, in case of cooking robots may need to observe a whole onion, take a knife and slice it. In this case, robots will grasp some object by any instrument with specific motion and perform the job. This specific motion is known as "functional motion". A number of robot grasping strategies \cite{lin2014grasp, lin2015robot} are already proposed. In \cite{lin2014grasp}, how robot grasping can gain knowledge from human grasping approach is discussed. A grasp synthesis technique is also proposed by observing the human grasping type, thumb placement and direction. In \cite{lin2015robot}, authors proposed a task-oriented grasp planning from the most significant part of a non-parametric statistical distribution model. In \cite{lin2012learning}, a Gaussian Mixer Model (GMM) based model is proposed to learn fingertip force for robot grasping which can be applied for manipulating any specific task. Any manipulation-oriented grasping needs proper wrench and motion to finish the task where the information is derived from the functionality of object states and instruments \cite{sun2016robotic}. Task Wrench Coverage and Task Manipulator Efficiency measures \cite{lin2015task} can lead us how good our grasping is for manipulating any task. Apart from grasping in our case, we also need proper knowledge of object relationship. In \cite{sun2014object}, authors proposed an object-object affordance learning approach using the Bayesian network to gain the knowledge of the object-object relationship. Recently, in \cite{paulius2016functional}, authors proposed a FOON network which establishes a relationship between the object states and the corresponding manipulated task. Robots can be learned about the object state and functional object motion by this FOON network. So here object state detection is very important to generate good knowledge by any network. 

In a kitchen environment robots may observe different cooking object states such as sliced, diced, grated, whole, creamy paste, julienne etc. Our goal is to detect these cooking state properly and provide a good knowledge of object states and manipulation to robots. 
Various kinds of image descriptors, specially pattern oriented descriptors \cite{2017POMF, 2017BanglaHandwritten} have been popular for feature extraction in computer vision related problems. But recently, different types of Convolutional Neural Networks (CNN) \cite{krizhevsky2012imagenet} are producing very good results in image analysis such as object detection. That is why we are also interested in analysis its impact on our cooking states problems. A lot of modified and ensemble versions of this neural network are already proposed such as AlexNet \cite{krizhevsky2012imagenet}, VGGNet (also called OxfordNet) \cite{simonyan2014very}, GoogleNet \cite{szegedy2015going}, Inception V3 \cite{szegedy2016rethinking}, Inception ResNet V2 \cite{szegedy2017inception} etc. Each model has its own architectural design, strength, and impact. In this paper, we exploit deep learning approach for identification of different cooking states from images using modified and fine-tuned model. We specifically focus on Inception V3 \cite{szegedy2016rethinking} and its impact on different cooking states. Finally, we propose an Inception V3 based model and test on recent Cooking State Recognition Challenge dataset \cite{2018arXiv180506956B} which achieves overall 75.65\% accuracy on validation dataset and 73.3\% on unseen test dataset. 

\section{Methodology}
We proposed a modified version of Inception V3 CNN networks in this research for cooking state identification problem. We fine-tuned the proposed model based on different common approaches of a deep neural network. The design of the proposed model is discussed in this section.

\subsection{Inception V3 CNN Model}
The development of Inception architecture started from the initial GoogleNet \cite{simonyan2014very} (also known as Inception V1). Google's researchers proposed that sparsity can be beneficial for the network. They introduced that based on Hebbian principle. Later on, researchers found that the convolutional structure larger than $(3*3)$ can be expressed more efficiently by factorizing to smaller convolutional layers. In \cite{szegedy2016rethinking}, they described some principles and optimization ideas as well as argued that it is possible to get the same layer effect more efficiently. 

\subsection{Proposed Model}
\begin{figure}
\centering
\fbox{\includegraphics[width=\linewidth]{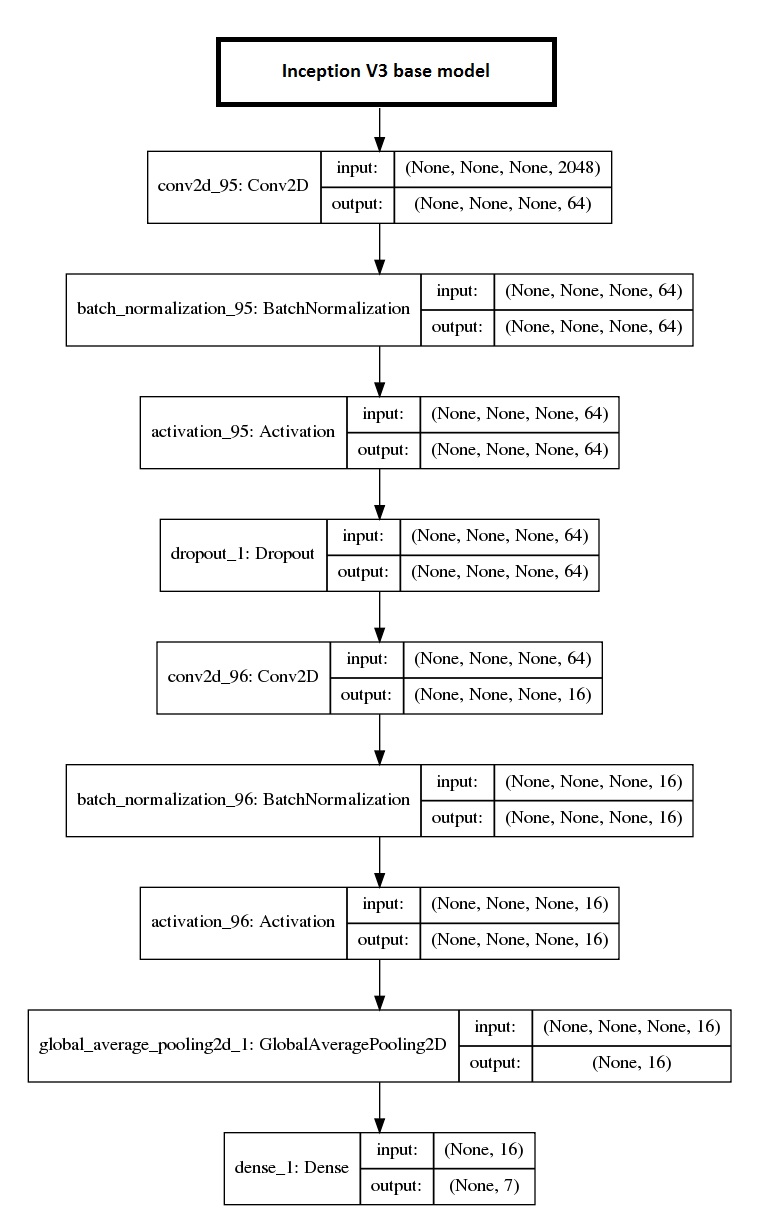}}
\caption{Proposed architecture.}
\label{fig3}
\end{figure}

In our proposed model, we add total two convolutional layers with a size of $64$, and $32$ as well as one fully connected dense layer with the original Inception V3 model. In all the convolutional layers, we use a $
(3*3)$ convolution mask to keep the computational cost minimal. As already Inception V3 builds a larger network, we wanted to keep the network computationally efficient. Besides the main purpose of Inception V3 CNN model was to avoid $(7*7)$ convolutional cost. Pooling layer usually reduces the size of the features. But in our modified model, we do not use any Max pooling layers at the top of the Inception V3 model because our input size is already small enough before reaching the modified layers. But we add one Global average pooling \cite{lin2013network} layer between the last two layers. It is shown that Global average pooling \cite{lin2013network} performs better in the last fully connected layer just before the softmax layer which is now being used as the alternate of dropout at the last feature extraction layer too. It makes a good relationship between feature map as categorical whereas traditional fully connected layers are dependent on dropout \cite{srivastava2014dropout} regularization. So we add this as like the original Inception V3 \cite{szegedy2016rethinking} CNN model. Besides, we also use Batch Normalization \cite{ioffe2015batch} after the convolution layers which forcefully makes a good initialization based on Gaussian distribution instead of random initialization at the beginning. The network of our proposed model is illustrated in Fig. \ref{fig3}. We applied regularization to our model using early stop and dropout approach. To optimize the model we try SGD, RMSprop, and Adam \cite{kingma2014adam} to find out the best optimization technique for our approach. However, SGD is found so far the best one to fit with our model and dataset.

\section{Evaluation and Results}
\subsection{Dataset and Implementation}
\begin{figure*}[!htb]
\includegraphics[width=\linewidth]{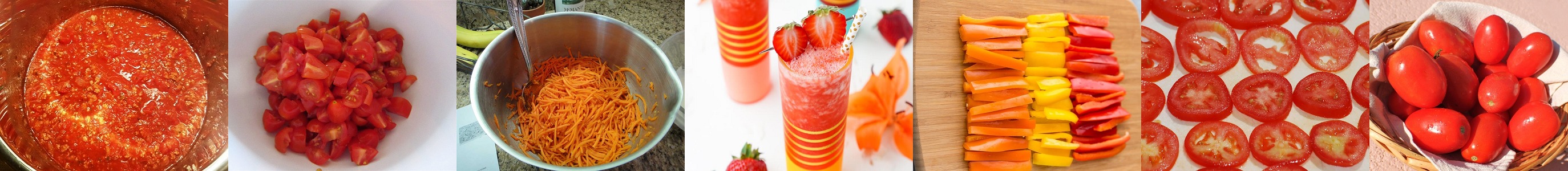}
\caption{Different Cooking States, from left to right: creamy paste, diced, grated, juiced, julienne, sliced, whole respectively.}
\label{data_preview}
\end{figure*}

In our experiments, we use the Cooking State Recognition Challenge dataset version 1.0 provided by \cite{2018arXiv180506956B}. It has total 5978 samples including 7 cooking states of around 18 types of objects. These are diced (700), julienne (672), sliced (1315), grated (819), whole (1304), juiced (638), creamy paste (730). In Fig. \ref{data_preview}, some of the samples of the dataset are shown. We implemented the proposed model using python, tensorflow, and keras.

\subsection{Data Preprocessing}
\begin{figure}
	\begin{multicols}{4}
		\includegraphics[width=\linewidth]{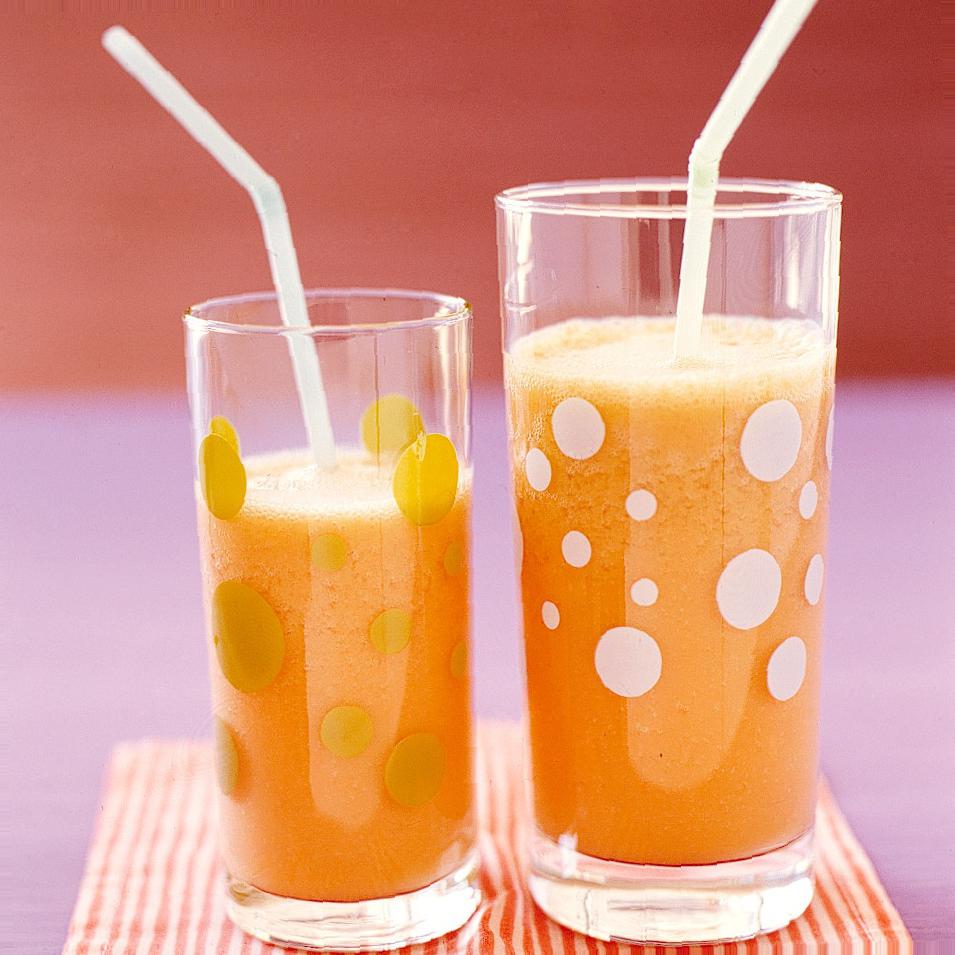}\par\caption*{(a)}
		\includegraphics[width=\linewidth]{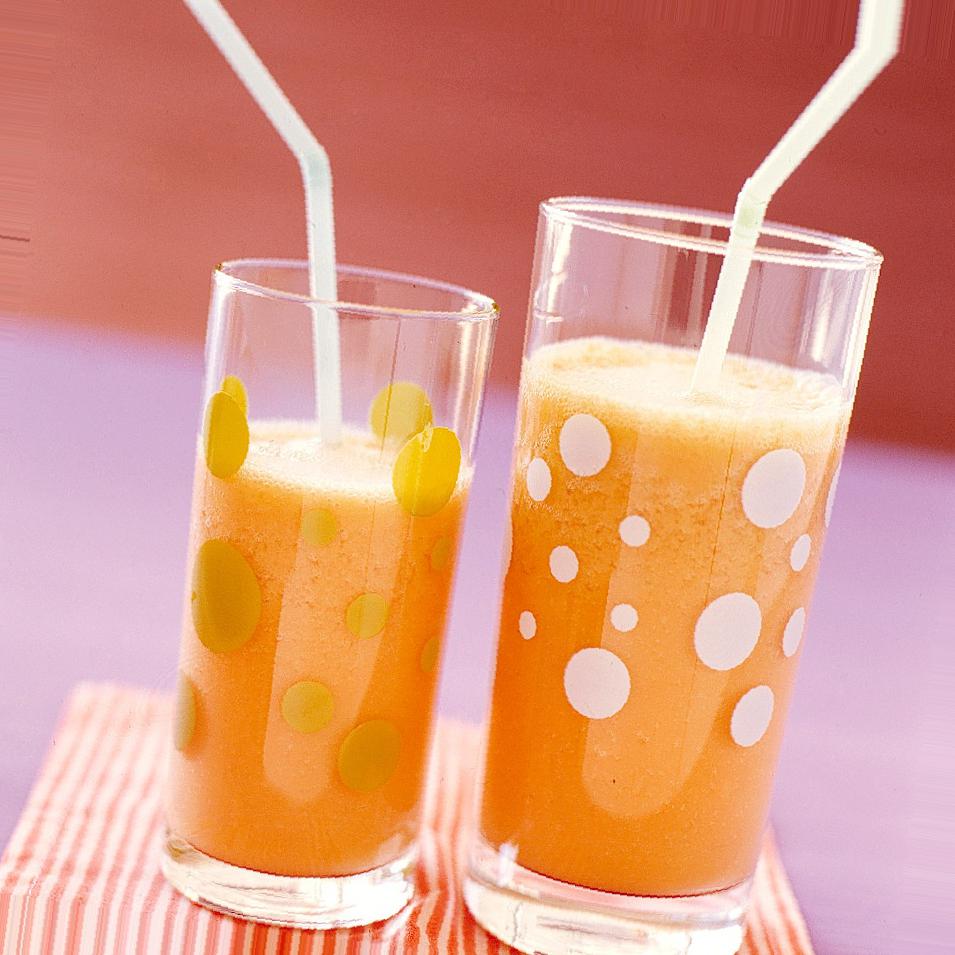}\par\caption*{(b)}
		\includegraphics[width=\linewidth]{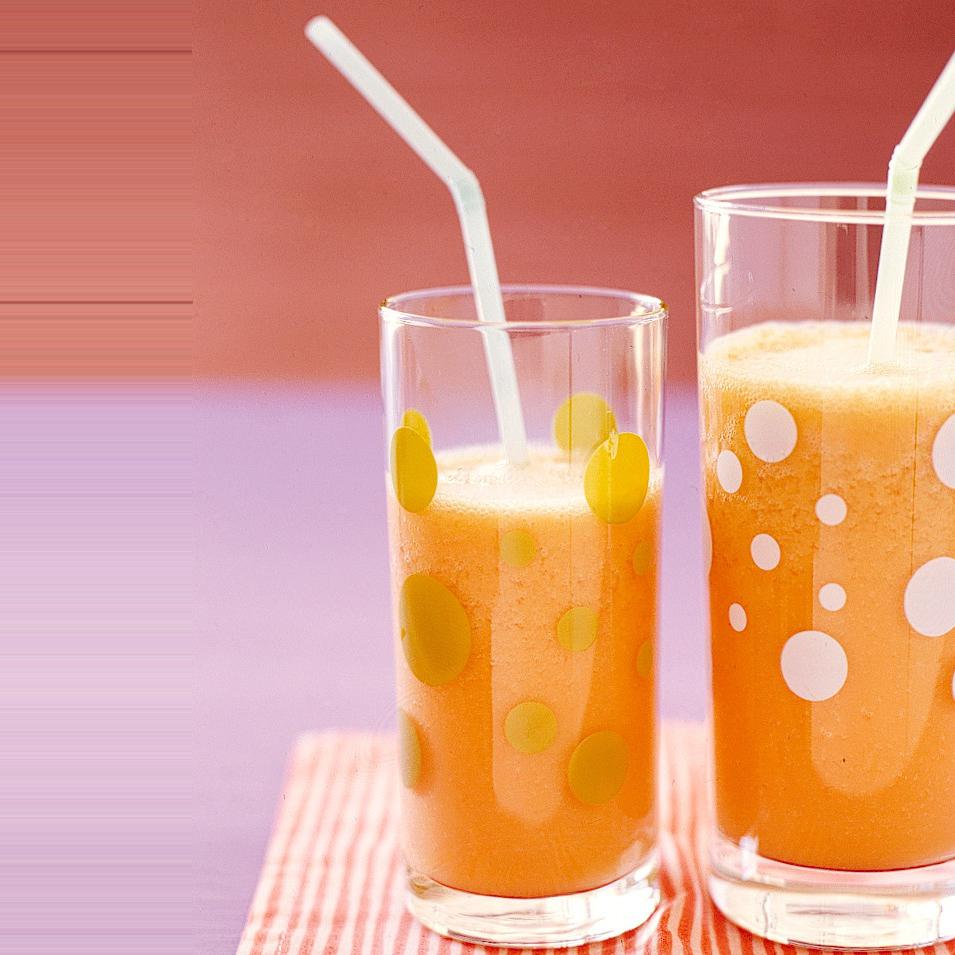}\par\caption*{(c)}
        \includegraphics[width=\linewidth]{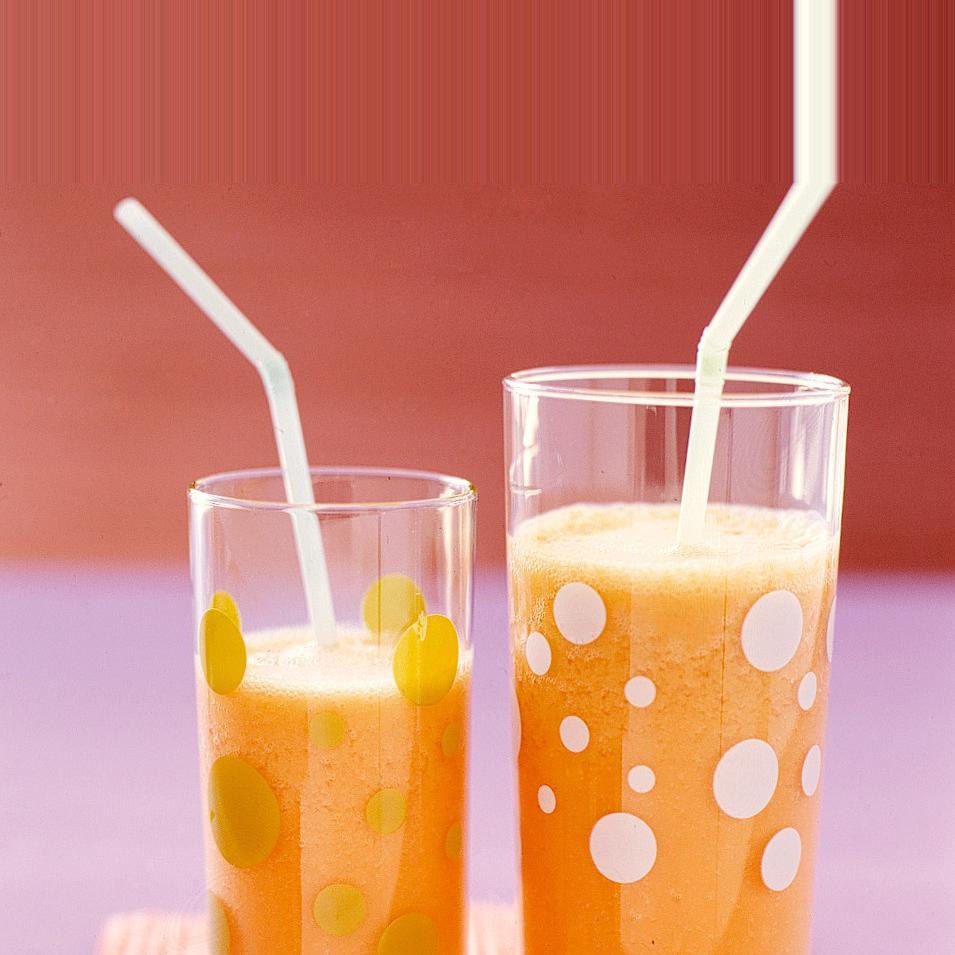}\par\caption*{(d)}
	\end{multicols}
\caption{Data augmentation examples: (a) original, (b) rotation, (c) width shift, (d) height shift.}
\label{fig1}
\end{figure}

Initially, we did some data preprocessing such as sample-wise centering and normalization. Sample-wise centering changes the image mean to $0$ and Sample-wise normalization divides the image by its standard deviation. Basically, Sample-wise mean transfers the geometric center of the image around the origin along every dimension. In our case, different images may have different scales so we normalized the image to get equal importance to the learning algorithm. Besides, we also make sure that all the image pixel values are distributed within the range of $0-255$ and resized to a fixed size. As in original Inception V3, it uses $299*299$ size. So we stick with that size in all of our experiments.
Apart from these we also perform data augmentation to make the data more robust and generalized. Basically, in data augmentation training data are altered and trained through the network to expose more variety of samples to the network. It helps the network to be trained in a robust and generalized way. As our dataset are not quite large enough for deep network, so proper augmentation plays a great role here. Augmentation involves image rotation, flipping, cropping, shearing etc. In our experiments, we created images considering up to $90$-degree rotation, 30\% horizontal and vertical shift, 30\% horizontal flip, 30\% shearing, 30\% zoom range. We did these because we believe in our dataset these may be the most common cases. In Fig. \ref{fig1}, examples are shown from the augmented image dataset.

\subsection{Model Training}
For our experiments, we randomly divide the entire dataset into two parts: 85\% for Training (5117) and 15\% for Testing (861). Our Training set of 5117 samples are again divided randomly into two parts: 80\% for Training (4124) and 20\% for Validation (994) which include all the 7 classes (diced, julienne, sliced, grated, whole, juiced, creamy paste). We have tried to analyze the results in terms of different batch size, optimization techniques. Apart from this, we also tried to figure out the layer effect for fine tuning of our designed model. Each of our experiments is done with an epoch size of 50. To avoid over-fitting, we did an early stopping observing the loss with a patience of 5 epochs during training. And during this training, we select the best weights for our model based on minimum validation loss.

\subsection{Model Evaluation}
We evaluated the proposed model based in terms of different parameters. We tried to figure out what are the effects of the model and how it can be efficiently fitted and trained for our cooking states problem. At the beginning of our training, we make sure that we are getting proper loss from the classifier. As softmax function is the final classifier so in our case the validation loss should be less than $-ln(0.14) = 1.97$ (total 7 cooking states). It reveals that our initialization is alright. In all our training loss and validation loss, we can observe this phenomenon. 


\subsubsection{Based on optimizers}
\begin{figure*}
	\begin{multicols}{3}
		\includegraphics[width=\linewidth]{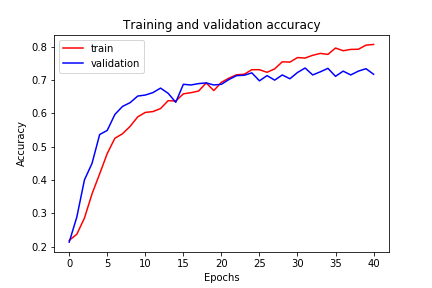}\par\caption*{(a) SGD batch size 32}
		\includegraphics[width=\linewidth]{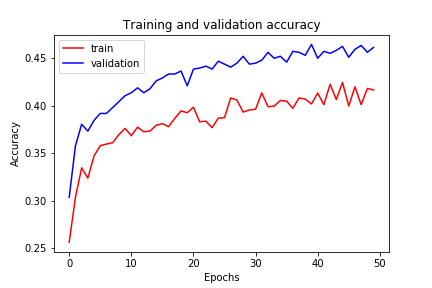}\par\caption*{(b) RMSprop batch size 32}
		\includegraphics[width=\linewidth]{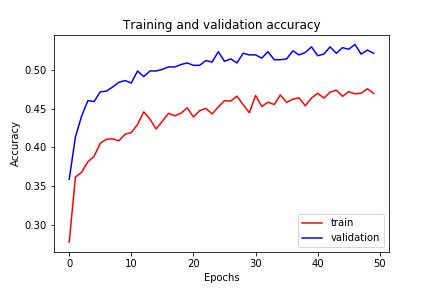}\par\caption*{(c) Adam batch size 32}
	\end{multicols}
	\begin{multicols}{3}
		\includegraphics[width=\linewidth]{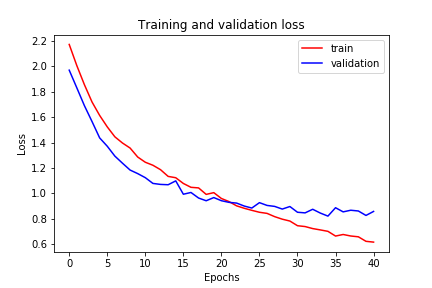}\par\caption*{(d) SGD batch size 32}
		\includegraphics[width=\linewidth]{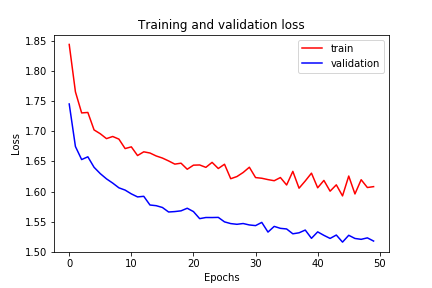}\par\caption*{(e) RMSprop batch size 32}
		\includegraphics[width=\linewidth]{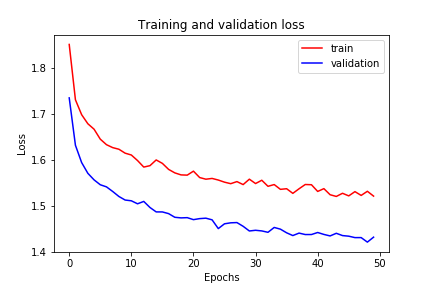}\par\caption*{(f) Adam batch size 32}
	\end{multicols}
\caption{Validation accuracy and loss during training for different optimizers.}
\label{fig5}
\end{figure*}

In our experiment, at first, we tried to find out the best optimizer for our model. We used SGD, RMSprop and Adam \cite{kingma2014adam}. Experimentally, we found SGD is superior to the others. Generally, for a small dataset, a higher learning rate suits best. For SGD empirically we selected the best learning rate for our model 0.001 and used a step decay. We found that step decay performs best for our model. Besides, we also found that using momentum 0.6 with an active Nesterov’s Accelerated Momentum shows best results. In case of RMSprop and Adam, we used the recommended setting from the original papers. In table \ref{tab1}, the comparative validation loss and accuracy for these optimizers without freezing any layers are shown. For all the cases, we used our default setting 50 epochs and early stopping with patience 5 epochs. Details of the progress during the training periods can also be found in the Fig. \ref{fig5}. We can observe that we reached at the highest validation accuracy of 75.65\% for SGD, whereas RMSprop and Adam both showed progressive but very lower accuracy for 50 epochs. But these may produce good accuracy with a higher number of epochs as well.

\begin{table*}
    \caption{Effect of different optimizers}
	\label{tab1}
	\centering
	\begin{tabular}{ |l|c|c| } 
		\hline
		Optimizer and parameter setup (Batch size 32 + 50 epochs + Dropout (50\%) + Early stopping + Best weight) & Val Loss & Val Accuracy \\
		\hline
		\hline
		SGD (LR = 0.001, step decay, momentum = 0.6, nesterov = True) & 0.7742 & 0.7565 \\
		\hline 
		RMSprop (LR = 0.001, decay = 0.999, rho = 0.9) & 1.5264 & 0.4557 \\ 
		\hline
		Adam (LR = 0.001, decay = 0.999, b1 = 0.9, b2 = 0.999, epsilon = 1e-8) & 1.4270 & 0.5271\\
		\hline
	\end{tabular}

\end{table*}

\subsubsection{Based on layers}
\begin{figure*}
	\begin{multicols}{3}
		\includegraphics[width=\linewidth]{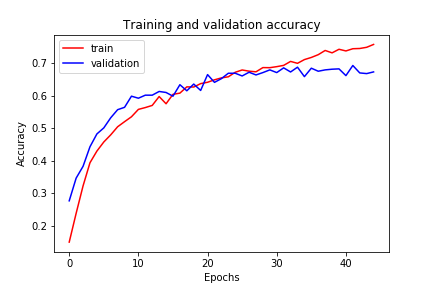}\par\caption*{(a) Freezing layer 100}
		\includegraphics[width=\linewidth]{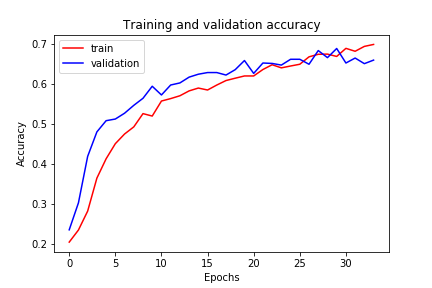}\par\caption*{(b) Freezing layer 132}
		\includegraphics[width=\linewidth]{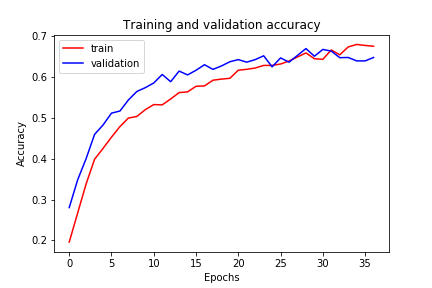}\par\caption*{(c) Freezing layer 164}
	\end{multicols}

	\begin{multicols}{3}
		\includegraphics[width=\linewidth]{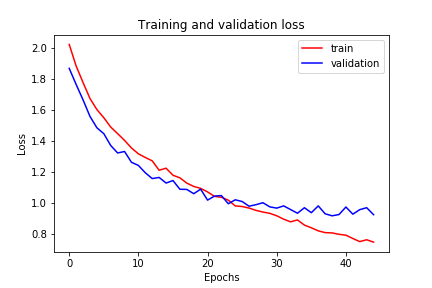}\par\caption*{(d) Freezing layer 100}
		\includegraphics[width=\linewidth]{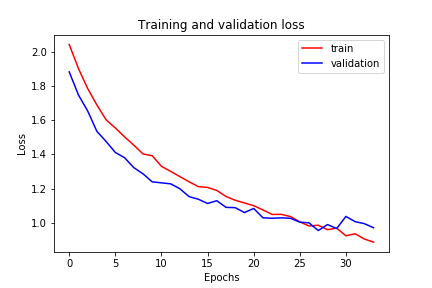}\par\caption*{(e) Freezing layer 132}
		\includegraphics[width=\linewidth]{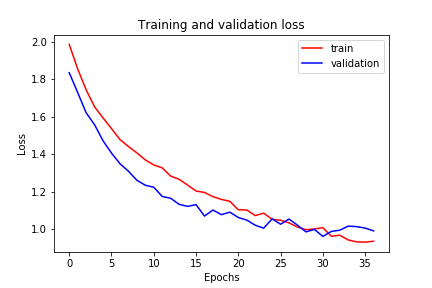}\par\caption*{(f) Freezing layer 164}
	\end{multicols}
\caption{Validation accuracy and loss during training for different freezing layers.}
\label{fig6}
\end{figure*}

We fine-tuned the proposed model by freezing number of layers. Our target was to find out up to which layer we can use the pre-trained model without training the weights with our dataset. In the original Inception V3, after a while, there were connected mixed layers from the previous sub-layers. As most of the image edges and shapes are trained by the beginning layers, so we froze up to mixed layer 4, 5 and 6 to see the effects. We have found that if we freeze more layers then the accuracy starts to decrease. If we do not freeze any layers then it shows the best performance for our training dataset. Although our trainable parameters increase if we start training from the beginning. The comparative analysis can be found in table \ref{tab2} and Fig. \ref{fig6}. It shows the loss starts to increase and accuracy starts to decrease with the increase of layer freezing. We believe as the pre-trained model were trained focusing on object detection but now our target is to detect object states, so if we train it from the beginning with our dataset it will produce better results.

\begin{table}
	\caption{Effect of different layers}
	\label{tab2}
	\centering
	\begin{tabular}{ |c|c|c|c|c| } 
		\hline
		Freezing & Total & Trainable & Validation & Validation \\
		Layers & Param. & Param. & Loss & Accuracy\\
		\hline
		\hline
		0-164 & 22,992,167 & 17,830,439 & 0.9951 & 0.6559 \\
		\hline
		0-132 & 22,992,167 & 19,519,719 & 0.9927 & 0.6670\\
		\hline 
		0-100 & 22,992,167 & 20,815,591 & 0.9480 & 0.6700 \\
		\hline
		No freeze   & 22,992,167 & 22,957,575 & 0.7742 & 0.7565 \\
		\hline
	\end{tabular}
\end{table}

\subsubsection{Based on batch size}
\begin{figure*}
	\begin{multicols}{3}
		\includegraphics[width=\linewidth]{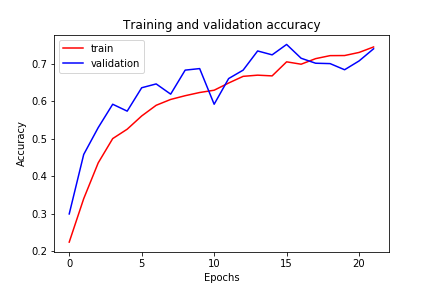}\par\caption*{(a) SGD batch size 16}
		\includegraphics[width=\linewidth]{figure/accuracy_SGD32.png}\par\caption*{(b) SGD batch size 32}
		\includegraphics[width=\linewidth]{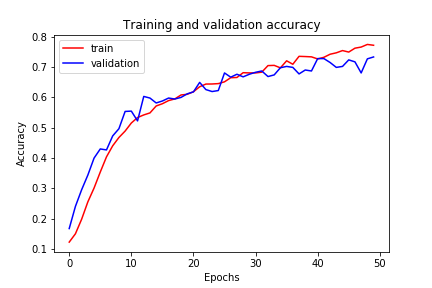}\par\caption*{(c) SGD batch size 64}
	\end{multicols}

	\begin{multicols}{3}
		\includegraphics[width=\linewidth]{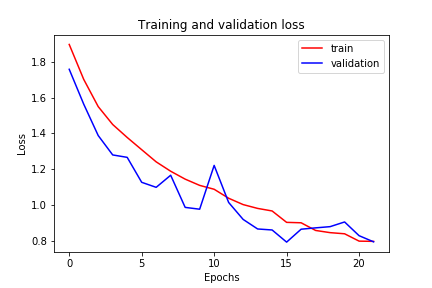}\par\caption*{(d) SGD batch size 16}
		\includegraphics[width=\linewidth]{figure/loss_SGD32.png}\par\caption*{(e) SGD batch size 32}
		\includegraphics[width=\linewidth]{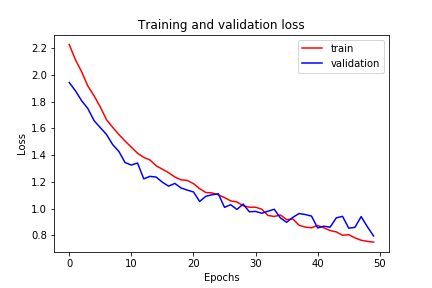}\par\caption*{(f) SGD batch size 64}
	\end{multicols}
\caption{Validation accuracy and loss during training for different batch size.}  
\label{fig7}
\end{figure*}

While the model was compiled, experimentally we observed that batch size 32 is an optimal value for our model using this dataset. If our batch size is too high or too small then we get a negative impact on the performance. Comparative performance using batch size 16, 32, 64 are shown in the table \ref{tab3}. Besides the performance of the validation loss and accuracy are also shown in the Fig. \ref{fig7}. We figure out that in our case batch size 32 shows the best performance. Both batch size 16 and 64 decrease the performance of the model in terms of validation accuracy and loss.

\begin{table}
	\caption{Effect of different batch sizes}
	\label{tab3}
	\centering
	\begin{tabular}{ |l|c|c| } 
		\hline
		Batch Size & Val Loss & Val Accuracy \\
		\hline
		\hline
		SGD (Batch 16) & 0.8241 & 0.7384 \\
		\hline 
		SGD (Batch 32) & 0.7742 & 0.7565 \\
		\hline
		SGD (Batch 64) & 0.8364 & 0.7293 \\
		\hline
	\end{tabular}
\end{table}

\subsubsection{Confusion matrix and Classification report}
\begin{figure}
		\includegraphics[width=\linewidth]{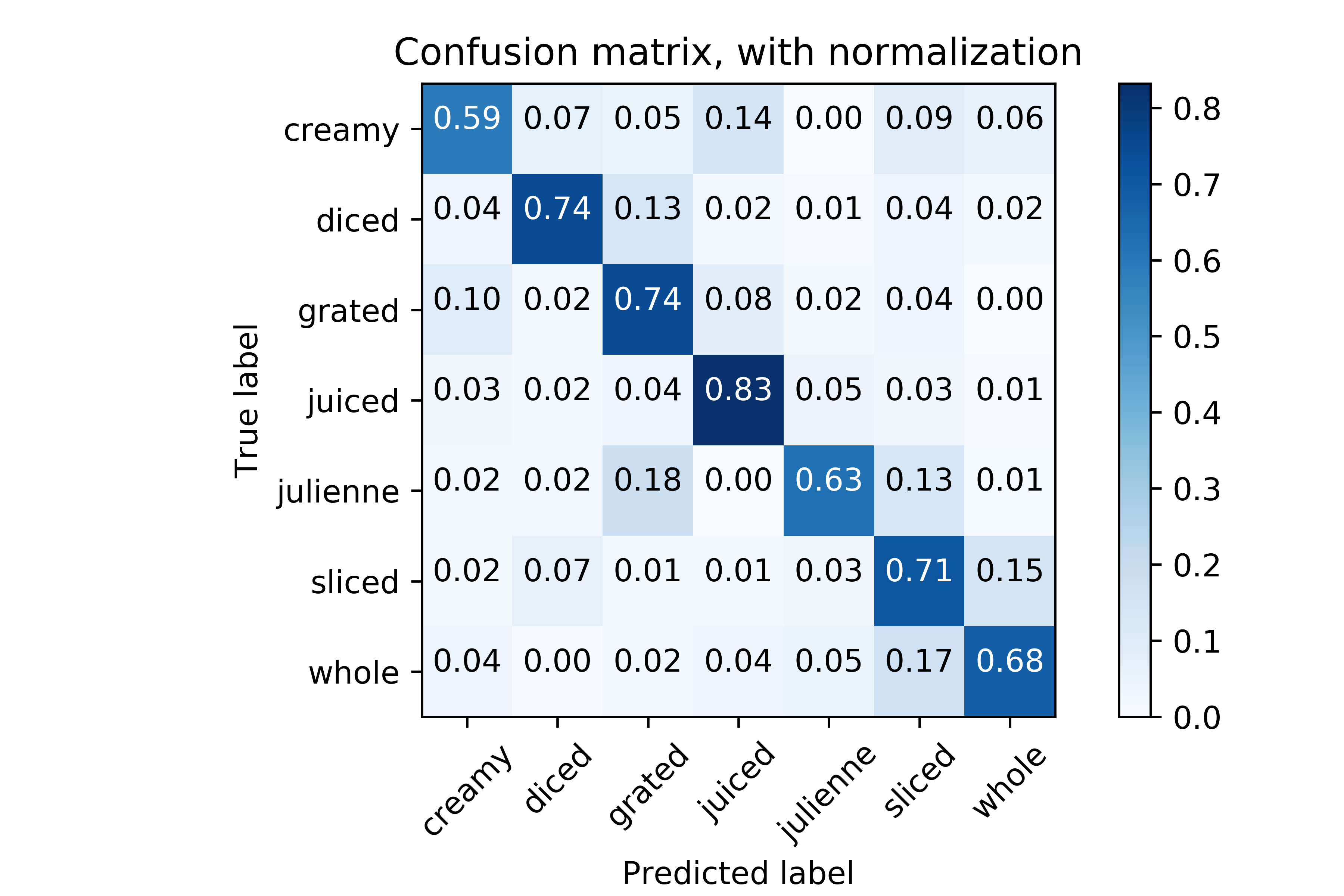}
	\caption{Normalized confusion matrix on validation dataset.}
    \label{fig8_2}
\end{figure}

\begin{figure}
		\includegraphics[width=\linewidth]{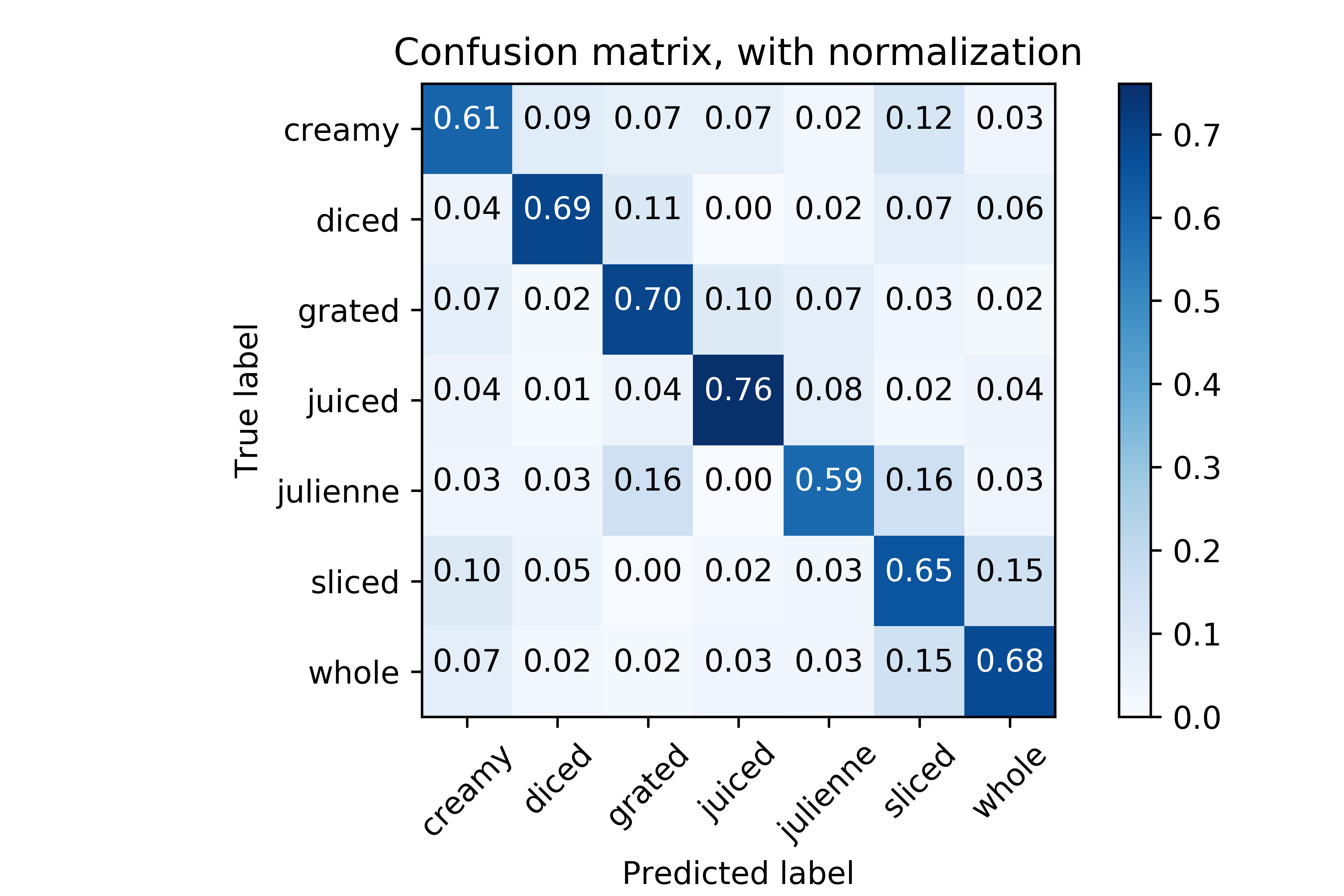}
	\caption{Normalized confusion matrix on testing dataset.}
    \label{fig9_2}
\end{figure}

Confusion matrix and classification report details (precision, recall, F-1 measure) reveal the overall summary of classification performance based on class by class. We generated these for both validation set and testing set. We got an overall 75.65\% accuracy on our validation dataset. In Fig. \ref{fig8_2}, and \ref{fig9_2}, normalized confusion matrices(\%) are shown. 

It can be observed that highest accuracy 83\% is found in case of juice and lowest accuracy 59\% in case of creamy. All others are within 70-80\% range. In table \ref{tab4}, the classification report are presented. Our precision, recall and F-1 score were 0.71, 0.70, and 0.70 respectively. In case of the testing set, we achieved 73.3\% accuracy and observed the almost same pattern of classifications. Juice was the best one (76\%), but this time both creamy (61\%) and julienne (59\%) got the lowest accuracy. It looks like that the most challenging one was the Creamy paste and Julienne because of a lot of variation of states which can be easily observable from the dataset. Juice had a structured state that is why its accuracy is the best. The details are shown in Fig. \ref{fig8_2} and \ref{fig9_2}. If we closely observe the normalized confusion matrix on both validation and test dataset we can get an overall idea about the data variation of the classes. We believe that if our training set would include all kinds of possible object states then it could produce more good results.
\begin{table}
	\caption{Classification report on validation dataset}
	\label{tab4}
	\centering
	\begin{tabular}{ |l|c|c|c|c| } 
		\hline
		\textbf{Class} & \textbf{Precision} & \textbf{Recall} & \textbf{F-1 score} & \textbf{Support}\\
        \hline
		\hline
		creamy & 0.66 & 0.59 & 0.62 & 117\\
        \hline
        diced & 0.75 & 0.74 & 0.75 & 128\\
        \hline
        grated & 0.67 & 0.74 & 0.70 & 132\\
        \hline
        juiced & 0.68 & 0.83 & 0.75 & 107\\
        \hline
        julienne & 0.68 & 0.63 & 0.65 & 83\\
        \hline
        sliced & 0.70 & 0.71 & 0.70 & 224\\
        \hline	
        whole & 0.76 & 0.68 & 0.72 & 203\\
        \hline
        average & 0.71 & 0.70 & 0.70 & 994\\
        \hline
	\end{tabular}
\end{table}

\section{Conclusion}
Cooking State Recognition is one of the main challenges for a kitchen robot. A robot needs to perceive clear idea about different cooking states (such as whether it is sliced or diced) to continue its cooking. In this research, we design a modified architecture  based on Inception CNN deep model, fine-tune its weights to detect different cooking states. We analyze the effect of different layers of the proposed deep model in terms of different parameters. Experimentally, it is found that proposed designed model can detect different cooking states with a significant accuracy of 75.65\% on the validation set and 73.3\% on the unseen test set of cooking state recognition challenge dataset. In future, we aim to do more experiment on relevant ImageNet dataset and the latest Cooking State Recognition dataset which are more robust.
 
\bibliographystyle{IEEEtran}
\bibliography{bib_database}

\begin{thebibliography}{10}
\providecommand{\url}[1]{#1}
\csname url@samestyle\endcsname
\providecommand{\newblock}{\relax}
\providecommand{\bibinfo}[2]{#2}
\providecommand{\BIBentrySTDinterwordspacing}{\spaceskip=0pt\relax}
\providecommand{\BIBentryALTinterwordstretchfactor}{4}
\providecommand{\BIBentryALTinterwordspacing}{\spaceskip=\fontdimen2\font plus
\BIBentryALTinterwordstretchfactor\fontdimen3\font minus
  \fontdimen4\font\relax}
\providecommand{\BIBforeignlanguage}[2]{{%
\expandafter\ifx\csname l@#1\endcsname\relax
\typeout{** WARNING: IEEEtran.bst: No hyphenation pattern has been}%
\typeout{** loaded for the language `#1'. Using the pattern for}%
\typeout{** the default language instead.}%
\else
\language=\csname l@#1\endcsname
\fi
#2}}
\providecommand{\BIBdecl}{\relax}
\BIBdecl

\bibitem{lin2014grasp}
Y.~Lin and Y.~Sun, ``Grasp planning based on strategy extracted from
  demonstration,'' in \emph{Intelligent Robots and Systems (IROS 2014), 2014
  IEEE/RSJ International Conference on}.\hskip 1em plus 0.5em minus 0.4em\relax
  IEEE, 2014, pp. 4458--4463.

\bibitem{lin2015robot}
Y.~Lin and Y.~\vspace{0mm}Sun, ``Robot grasp planning based on demonstrated
  grasp strategies,'' \emph{The International Journal of Robotics Research},
  vol.~34, no.~1, pp. 26--42, 2015.

\bibitem{lin2012learning}
Y.~Lin, S.~Ren, M.~Clevenger, and Y.~Sun, ``Learning grasping force from
  demonstration,'' in \emph{Robotics and Automation (ICRA), 2012 IEEE
  International Conference on}.\hskip 1em plus 0.5em minus 0.4em\relax IEEE,
  2012, pp. 1526--1531.

\bibitem{sun2016robotic}
Y.~Sun, Y.~Lin, and Y.~Huang, ``Robotic grasping for instrument
  manipulations,'' in \emph{Ubiquitous Robots and Ambient Intelligence (URAI),
  2016 13th International Conference on}.\hskip 1em plus 0.5em minus
  0.4em\relax IEEE, 2016, pp. 302--304.

\bibitem{lin2015task}
Y.~Lin and Y.~Sun, ``Task-based grasp quality measures for grasp synthesis,''
  in \emph{Intelligent Robots and Systems (IROS), 2015 IEEE/RSJ International
  Conference on}.\hskip 1em plus 0.5em minus 0.4em\relax IEEE, 2015, pp.
  485--490.

\bibitem{sun2014object}
Y.~Sun, S.~Ren, and Y.~Lin, ``Object--object interaction affordance learning,''
  \emph{Robotics and Autonomous Systems}, vol.~62, no.~4, pp. 487--496, 2014.

\bibitem{paulius2016functional}
D.~Paulius, Y.~Huang, R.~Milton, W.~D. Buchanan, J.~Sam, and Y.~Sun,
  ``Functional object-oriented network for manipulation learning,'' in
  \emph{Intelligent Robots and Systems (IROS), 2016 IEEE/RSJ International
  Conference on}.\hskip 1em plus 0.5em minus 0.4em\relax IEEE, 2016, pp.
  2655--2662.

\bibitem{2017POMF}
M.~H. Kabir, M.~S. Salekin, M.~Z. Uddin, and M.~Abdullah-Al-Wadud, ``Facial
  expression recognition from depth video with patterns of oriented motion
  flow,'' \emph{IEEE Access}, vol.~5, pp. 8880--8889, 2017.

\bibitem{2017BanglaHandwritten}
T.~I. Aziz, A.~S. Rubel, M.~S. Salekin, and R.~Kushol, ``Bangla handwritten
  numeral character recognition using directional pattern,'' in \emph{2017 20th
  International Conference of Computer and Information Technology (ICCIT)}, Dec
  2017, pp. 1--5.

\bibitem{krizhevsky2012imagenet}
A.~Krizhevsky, I.~Sutskever, and G.~E. Hinton, ``Imagenet classification with
  deep convolutional neural networks,'' in \emph{Advances in neural information
  processing systems}, 2012, pp. 1097--1105.

\bibitem{simonyan2014very}
K.~Simonyan and A.~Zisserman, ``Very deep convolutional networks for
  large-scale image recognition,'' \emph{arXiv preprint arXiv:1409.1556}, 2014.

\bibitem{szegedy2015going}
C.~Szegedy, W.~Liu, Y.~Jia, P.~Sermanet, S.~Reed, D.~Anguelov, D.~Erhan,
  V.~Vanhoucke, A.~Rabinovich \emph{et~al.}, ``Going deeper with
  convolutions.''\hskip 1em plus 0.5em minus 0.4em\relax Cvpr, 2015.

\bibitem{szegedy2016rethinking}
C.~Szegedy, V.~Vanhoucke, S.~Ioffe, J.~Shlens, and Z.~Wojna, ``Rethinking the
  inception architecture for computer vision,'' in \emph{Proceedings of the
  IEEE Conference on Computer Vision and Pattern Recognition}, 2016, pp.
  2818--2826.

\bibitem{szegedy2017inception}
C.~Szegedy, S.~Ioffe, V.~Vanhoucke, and A.~A. Alemi, ``Inception-v4,
  inception-resnet and the impact of residual connections on learning.'' in
  \emph{AAAI}, vol.~4, 2017, p.~12.

\bibitem{2018arXiv180506956B}
A.~B. {Jelodar}, M.~S. {Salekin}, and Y.~{Sun}, ``Identifying object states in
  cooking-related images,'' \emph{arXiv preprint arXiv:1805.06956}, May 2018.

\bibitem{lin2013network}
M.~Lin, Q.~Chen, and S.~Yan, ``Network in network,'' \emph{arXiv preprint
  arXiv:1312.4400}, 2013.

\bibitem{srivastava2014dropout}
N.~Srivastava, G.~Hinton, A.~Krizhevsky, I.~Sutskever, and R.~Salakhutdinov,
  ``Dropout: A simple way to prevent neural networks from overfitting,''
  \emph{The Journal of Machine Learning Research}, vol.~15, no.~1, pp.
  1929--1958, 2014.

\bibitem{ioffe2015batch}
S.~Ioffe and C.~Szegedy, ``Batch normalization: Accelerating deep network
  training by reducing internal covariate shift,'' in \emph{International
  conference on machine learning}, 2015, pp. 448--456.

\bibitem{kingma2014adam}
D.~P. Kingma and J.~Ba, ``Adam: A method for stochastic optimization,''
  \emph{arXiv preprint arXiv:1412.6980}, 2014.

\end{thebibliography}

\end{document}